\def\year{2022}\relax
\documentclass[letterpaper]{article} 
\usepackage[english]{babel}
\usepackage{lmodern}
\usepackage{amssymb}
\usepackage{amsmath}
\usepackage{aaai22}  
\usepackage{times}  
\usepackage{helvet}  
\usepackage{courier}  
\usepackage[hyphens]{url}  
\usepackage{graphicx} 
\usepackage{natbib}  
\usepackage{caption} 
\DeclareCaptionStyle{ruled}{labelfont=normalfont,labelsep=colon,strut=off} 
\usepackage{algorithm}
\usepackage{algorithmic}
\usepackage{subfigure}
\usepackage{multirow} 
\usepackage{multicol} 

\usepackage{newfloat}
\usepackage{listings}
\floatstyle{ruled}
\newfloat{listing}{tb}{lst}{}
\floatname{listing}{Listing}
\title{Deep Supervised Information Bottleneck Hashing for Cross-modal Retrieval based Computer-aided Diagnosis}
\author{
    Yufeng Shi\textsuperscript{\rm 1}, Shuhuang Chen\textsuperscript{\rm 1},
    Xinge You\textsuperscript{\rm 1\footnote{Contact Author}}, Qinmu Peng\textsuperscript{\rm 1}, Weihua Ou\textsuperscript{\rm 2}, Yue Zhao\textsuperscript{\rm 3}
}
\affiliations{
    \textsuperscript{\rm 1}Huazhong University of Science and Technology\\
    \textsuperscript{\rm 2}Guizhou Normal University\\
    \textsuperscript{\rm 3}Hubei University\\
    \{yufengshi17, shuhuangchen, youxg, pengqinmu\}@hust.edu.cn, ouweihuahust@gmail.com, zhaoyhu@hubu.edu.cn
}

\begin{document}

\maketitle

\begin{abstract}
Mapping X-ray images, radiology reports, and other medical data as binary codes in the common space, which can assist clinicians to retrieve pathology-related data from heterogeneous modalities~(i.e., hashing-based cross-modal medical data retrieval), provides a new view to promot computer-aided diagnosis. Nevertheless, there remains a barrier to boost medical retrieval accuracy: how to reveal the ambiguous semantics of medical data without the distraction of superfluous information. To circumvent this drawback, we propose Deep Supervised Information Bottleneck Hashing~(DSIBH), which effectively strengthens the discriminability of hash codes. Speciﬁcally, the Deep Deterministic Information Bottleneck~\cite{yu2021deep} for single modality is extended to the cross-modal scenario. Beneﬁting from this, the superfluous information is reduced, which facilitates the discriminability of hash codes. Experimental results demonstrate the superior accuracy of the proposed DSIBH compared with state-of-the-arts in cross-modal medical data retrieval tasks.



\end{abstract}

\noindent The rapid development of medical technology not only provides diverse medical examinations but also produces tremendous amounts of medical data, ranging from X-ray images to radiology reports. It is an experience-demanding, time-consuming, and error-prone job to manually assess medical data and diagnose disease. To reduce the work burden of physicians and optimize the diagnostic process, computer-aided diagnosis~(CAD) systems including classifier based CAD\cite{shi2020graph,INES2021105782} and content-based image retrieval~(CBIR) based CAD\cite{yang2020deep,fang2021deep} have been designed to automatically identify illness. Although the two types of methods greatly promote the development of CAD, existing systems ignore the character of current medical data, which is diverse in modality and huge in terms of scale. Therefore, we introduce cross-modal retrieval~(CMR)~\cite{wang2016comprehensive} techniques and construct a CMR-based CAD method using semantic hashing~\cite{wang2017survey} to handle the above challenges.

With the help of CMR that projects multimodal data into the common space, samples from different modalities can be directly matched without the interference of heterogeneity. Therefore, CMR-based CAD can not only retrieve the semantically similar clinical profiles in heterogeneous modalities but also provide diagnosis results according to the previous medical advice. Compared with the classifier-based CAD that only provides diagnosis results, CMR-based CAD is more acceptable due to the interpretability brought by retrieved profiles. Compared with the CBIR-based CAD, CMR-based CAD wins on its extended sight of multi-modal data, which meets the requirement of current medical data. 

Recently, extensive work on hashing-based CMR that maps data from different modalities into the same hamming space, has been done by researchers to achieve CMR~\cite{li2018self,zhu2020flexible,yus2021deep}. Due to its compact binary codes and XOR distance calculation, hashing-based CMR possesses low memory usage and high query speed~\cite{wang2017survey}, which is also compatible with the huge volume of current medical data. In terms of accuracy, the suitable hashing-based solutions for CMR-based CAD are deep supervised hashing~(DSH) methods~\cite{xie2020multi,10.1145/3394171.3413962,10.1145/3460426.3463625}. With the guidance of manual annotations, deep supervised methods usually perform hash code learning based on the original data via neural networks. Inspired by the information bottleneck principle~\cite{tishby1999information}, the above-mentioned optimization procedure can be viewed as building hash code $G$ about a semantic label $Y$ through samples in different modalities $X=\left\{ X^1,X^2 \right\}$, which can be formulated as:
\begin{align}
\max\mathcal{L}=I\left( G;Y \right) -\beta I\left( G;X  \right),\label{shi_eq1}
\end{align}
where $I\left( \cdot;\cdot \right)$ represents the mutual information, and $\beta$ is a hyper-parameter. As quantified by $I\left( G;Y \right)$, current DSH methods model the semantic annotations to establish pair-wise, triplet-wise or class-wise relations, and maximize the correlation between hash codes and the semantic relations. Despite the consideration of semantic relations, the neglect of $I\left( G; X \right)$ will result in the retention of redundant information in the original data, thus limiting the improvement of the retrieval accuracy. $I\left( G;X \right)$ measures the correlation between the hash code $G$ and the data from two modalities $X$, which can be used to reduce the superfluous information from medical data, and constrain the hash code to grasp the correct semantics from annotations. Therefore, it can be expected that the optimization of Eq.~(\ref{shi_eq1}) can strengthen the discriminability of hash codes, which improves the accuracy of CMR-based CAD.

To perform CMR-based CAD, we design a novel method named Deep Supervised Information Bottleneck Hashing~(DSIBH), which optimizes the information bottleneck to strengthen the discriminability of hash codes. Specifically, to avoid variational inference and distribution assumption, we extend the Deep Deterministic Information Bottleneck~(DDIB)~\cite{yu2021deep} from single modality to the cross-modal scenario for hash code learning.

To summarize, our main contributions are fourfold:
\begin{itemize}
\item The cross-modal retrieval technique based on semantic hashing is introduced to establish computer-aided diagnosis systems, which is suitable for the current large-scale multi-modal medical data.
\item A deep hashing method named DSIBH, which optimizes the hash code learning procedure following the information bottleneck principle to reduce the distraction of superfluous information, is proposed for CMR-based CAD. 
\item To reduce the adverse impact of variational inference or distribution assumption, the Deep Deterministic Information Bottleneck is elegantly extended to the cross-modal scenario for hash code learning.
\item Experiments on the large-scale multi-modal medical dataset MIMIC-CXR show that DSIBH can strengthen the discriminability of hash codes more effectively than other methods, thus boosting the retrieval accuracy.
\end{itemize}

\section{Related Work}\label{sec2}
In this section, representative CAD approaches and hashing-based solutions of cross-modal retrieval are briefly reviewed. To make readers easier to understand our work, some knowledge of the DDIB is also introduced.

\subsection{Computer-aided Diagnosis}

CAD approaches generally fall into two types including classifier-based CAD and CBIR-based CAD. Thanks to the rapid progress of deep learning, classifier-based CAD methods~\cite{zhang2019attention,de2020deep} can construct task-specific neural networks to categorize histopathology images and employ the outcomes as the diagnosis. On the other side, CBIR-based CAD can provide clinical evidence since they retrieve and visualize images with the most similar morphological profiles. According to the data type of representations, existing CBIR methods can be divided into continuous value CBIR~\cite{erfankhah2019heterogeneity,2020Deep} and hashing-based CBIR~\cite{2020Creating,yang2020deep}. In the age of big data, the latter increasingly become mainstream due to the low memory usage and high query speed brought by hashing. Although substantial efforts have been made to analyse clinical image, medical data such as radiology reports in other modalities are ignored. Consequently, CAD is restricted in single modality and the cross-modal relevance between different modalities still waits to be explored. 
\subsection{Cross-modal Retrieval}
Cross-modal hashing has made remarkable progress in handling cross-modal retrieval, and this kind of methods can be roughly divided into two major types including unsupervised approaches and supervised approaches in terms of the consideration of semantic information. Due to the absence of semantic information, the former usually relies on data distributions to align semantic similarities of different modalities~\cite{liu2020joint,yus2021deep}. For example, Collective Matrix Factorization Hashing~\cite{ding2016large} learns unified hash codes by collective matrix factorization with a latent factor model to capture instance-level correlations. Recently, Deep Graph-neighbor Coherence Preserving Network~\cite{yus2021deep} extra explores graph-neighbor coherence to describe the complex data relationships. Although data distributions indeed help to solve cross-modal retrieval to some extent, one should note that unsupervised methods fail to manage the high-level semantic relations due to the neglect of manual annotations.

Supervised hashing methods are thereafter proposed to perform hash code learning with the guidance of manual annotations. Data points are encoded to express semantic similarity such as pair-wise\cite{shen2017deep,wang2019fusion}, triplet-wise~\cite{hu2019triplet,song2021deep} or multi-wise similarity relations\cite{cao2017collective,li2018self}. As an early attempt with deep learning, Deep Cross-modal Hashing~\cite{jiang2017deep} directly encodes original data points by minimizing the negative log likelihood of the cross-modal similarities. To discover high-level semantic information, Self-Supervised Adversarial Hashing~\cite{li2018self} harnesses a self-supervised semantic network to preserve the pair-wise relationships. Although various relations have been built between the hash code and the semantic labels, the aforementioned algorithms still suffer from the distraction of superfluous information, which is caused by the connections between the hash code and the original data, 

Consequently, for CMR-based CAD, there remains a need for a deep hashing method which can reduce the superfluous information to strengthen the discriminability of hash codes. 

\subsection{Deep Deterministic Information Bottleneck}
Despite great efforts to handle the ambiguous semantics of medical data, the discriminability of hash codes still needs to be strengthened. To alleviate such limitation, a promising solution is Deep Deterministic Information Bottleneck~\cite{yus2021deep} that has been proved to reduce the superfluous information during feature extraction. Before elaborating on our solution, we introduce basic knowledge on DDIB below.

DDIB intends to adopt a neural network to parameterize information bottleneck~\cite{tishby1999information}, which considers extracting information about a target signal $Y$ through a correlated observable $X$. The extracted information is represented as a variable $T$. The information extraction process can be formulated as:
\begin{align}
\max\mathcal{L}_{IB}=I\left( T;Y \right) -\beta I\left( T;X \right).\label{shi_eq2}
\end{align}

When the above objective is optimized with a neural network, $T$ is the output of one hidden layer. To update the parameters of networks, the second item in Eq.~(\ref{shi_eq2}) is calculated with the differentiable matrix-based R$\acute{e}$nyi's $\alpha$-order mutual information:
\begin{align}
I_{\alpha}\left( X;T \right) =H_{\alpha}\left( X \right) +H_{\alpha}\left( T \right) -H_{\alpha}\left( X,T \right),\label{shi_eq3}
\end{align}
where $H_{\alpha}\left( \cdot \right)$ indicates the matrix-based analogue to R$\acute{e}$nyi's $\alpha$-entropy and $H_{\alpha}\left( \cdot,\cdot \right)$ is the matrix-based analogue to R$\acute{e}$nyi's $\alpha$-order joint-entropy. More details of the matrix-based R$\acute{e}$nyi's $\alpha$-order entropy functional can be found in ~\cite{yu2019multivariate}.

For the first item in Eq.~(\ref{shi_eq2}), since $I\left( T;Y \right) =H\left( Y \right) -H\left( Y|T \right) $, the maximization of $I\left( T;Y \right)$ is converted to the minimization of $H\left( Y|T \right)$. Given the training set $\left\{ x_i,y_i \right\} _{i=1}^{N}$, the average cross-entropy loss is adopted to minimize the $H\left( Y|T \right)$:
\begin{align}
\frac{1}{N}\sum_{i=1}^N{\mathbb{E}_{t\sim p\left( t|x_i \right)}\left[ -\log p\left( y_i|t \right) \right]},\label{yu_cross}
\end{align}

Therefore, DDIB indicates that the optimization of Information Bottleneck in single modality can be achieved with a cross-entropy loss and a differentiable mutual information item $I\left( T;X \right)$. Obviously, the differentiable optimization strategy of information bottleneck in DDIB can benefit DSH methods in terms of superfluous information reduction. 
\section{Method}\label{sec3}
In this section, we first present the problem definition, and then detail our DSIBH method. The optimization is finally given. For illustration purposes, our DSIBH is applied in  X-ray images and radiology reports.
\subsection{Notation and problem definition}\label{notation}
Matrix and vector used in this paper are represented by boldface uppercase letter~(e.g., $\bf{G}$) and boldface lowercase letter~(e.g., $\bf{g}$) respectively. $\left\|\cdot\right\|$ denotes the 2-norm of vectors. $sign(\cdot)$ is defined as the sign function, which outputs 1 if its input is positive else outputs -1.
	
Let $\boldsymbol{X}^{1}=\left\{ \boldsymbol{x}_{i}^{1} \right\} _{i=1}^N$ and $\boldsymbol{X}^{2}=\left\{ \boldsymbol{x}_{j}^{2} \right\} _{j=1}^N$ symbolize X-ray images and radiology reports in the training set, where $\boldsymbol{x}_{i}^{1}\in \mathbb{R}^{d_1}$, $\boldsymbol{x}_{j}^{2}\in \mathbb{R}^{d_2}$. Their semantic labels that indicate the existence of pathology are represented by $\boldsymbol{Y}=\left\{ \boldsymbol{y}_l \right\} _{l=1}^N$, where $\boldsymbol{y}_l=\left\{ y_{l1},y_{l{2,}}...,y_{ld_3} \right\} \in \mathbb{R}^{d_3}$. Following~\cite{cao2016correlation,jiang2017deep,li2018self}, we define the semantic affinities $\boldsymbol{S}_{N\times N}$ between $\boldsymbol{x}_{i}^{1}$ and $\boldsymbol{x}_{j}^{2}$ using semantic labels. If $\boldsymbol{x}_{i}^{1}$ and $\boldsymbol{x}_{j}^{2}$ share at least one category label, they are semantically similar and $\boldsymbol S_{ij}=1$. Otherwise, they are semantically dissimilar and thus $\boldsymbol S_{ij}=0$. 

The goal of the proposed DSIBH is to learn hash functions $f^{1}\left( \theta _{1};\boldsymbol{X}^{1} \right) :\mathbb{R}^{d_{1}}\rightarrow \mathbb{R}^{d_{c}}$ and $f^{2}\left( \theta _{2};\boldsymbol{X}^{2} \right) :\mathbb{R}^{d_{2}}\rightarrow \mathbb{R}^{d_{c}}$, which can map X-ray images and radiology reports as approximate binary codes $\boldsymbol{G}^1$ and $\boldsymbol{G}^{2}$ in the same continuous space respectively. Later, binary codes can be generated by applying a sign function to $\boldsymbol{G}^{1,2}$.

Meanwhile, hamming distance $D\left( \boldsymbol{g}_{i}^{1},\boldsymbol{g}_{j}^{2} \right) $ between hash codes $\boldsymbol{g}_{i}^{1}$ and $\boldsymbol{g}_{j}^{2}$ needs to indicate the semantic similarity $\boldsymbol S_{ij}$ between $\boldsymbol{x}_{i}^{1}$ and $\boldsymbol{x}_{j}^{2}$, which can be formulated as:
\begin{align}
\boldsymbol{S}_{ij}\propto -D\left( \boldsymbol{g}_{i}^{1},\boldsymbol{g}_{j}^{2}\right).
\label{eq5}\end{align}
	
\subsection{Information Bottleneck in Cross-modal Scenario}
To improve the accuracy of CMR-based CAD, the superfluous information from the medical data in the hash code learning procedure should be reduced via the information bottleneck principle. Therefore, the information bottleneck principle in single modality should be extended to the cross-modal scenario, where one instance can own descriptions in different modalities. 

Analysis starts from the hash code learning processes for X-ray images and radiology reports respectively. Following the information bottleneck principle, the basic objective functions can be formulated as:
\begin{align}
\max\mathcal{L}_{IB\_1}=I\left( G^1;Y^1 \right) -\beta I\left( G^1;X^1 \right)\nonumber, \\\max\mathcal{L}_{IB\_2}=I\left( G^2;Y^2 \right) -\beta I\left( G^2;X^2 \right).
\label{shi_eq3}\end{align}
In cross-modal scenario, X-ray images and radiology reports in the training set are collected to describe the common pathology. Therefore, the image-report pairs own the same semantic label. To implement this idea, Eq.~(\ref{shi_eq3}) is transformed to:
\begin{align}
\max\mathcal{L}_{IB\_1}=I\left( G^1;Y \right) -\beta I\left( G^1;X^1 \right)\nonumber, \\\max\mathcal{L}_{IB\_2}=I\left( G^2;Y \right) -\beta I\left( G^2;X^2 \right),
\label{shi_eq4}\end{align}
Furthermore, the same hash code should be assigned for the paired samples to guarantee the consistency among different modalities, which is achieved with the $\ell _2$ loss:
\begin{align}
\min\mathcal{L}_{CONS}=\mathbb{E}\left[ \left\| g^1-g^y \right\| ^2 \right] +\mathbb{E}\left[ \left\| g^2-g^y \right\| ^2 \right],
\label{shi_eq5}\end{align}
where $G^y$ represents the modality-invariant hash codes for the image-report pairs. 

Incorporating Eq.~(\ref{shi_eq4}) and Eq.~(\ref{shi_eq5}), the overall
objective of the information bottleneck principle in cross-modal scenario is formulated as:
\begin{align}\label{shi_eq6}
\max\mathcal{L}_{IB\_C}=&\left( I\left( G^1;Y \right)+I\left( G^2;Y \right) \right)\\\nonumber&-\beta \left( I\left( G^1;X^1 \right) +I\left( G^2;X^2 \right) \right) \\\nonumber&-\gamma \left( \mathbb{E}\left[ \left\| g^1-g^y \right\| ^2 \right] +\mathbb{E}\left[ \left\| g^2-g^y \right\| ^2 \right] \right).
\end{align}

\subsection{Deep Supervised Information Bottleneck Hashing}
Following the information bottleneck principle in cross-modal scenario~(i.e., Eq.~(\ref{shi_eq6})), three variables including $G^y$, $G^1$ and $G^2$ should be optimized. To obtain modality-invariant $G^y$, we build labNet $f^y$ to directly transform semantic labels into the pair-level hash codes. The labNet is formed by a two-layer Multi-Layer Perception~(MLP) whose nodes are 4096 and $c$. Then, we build imgNet $f^1$ and txtNet $f^2$ as hash functions to generate hash codes $G^1$ and $G^2$. For X-ray images, we modify CNN-F~\cite{chatfield2014return} to build imgNet with the consideration of network scale. To obtain $c$ bit length hash codes, the last fully-connected layer in the origin CNN-F is changed to a $c$-node fully-connected layer. For radiology reports, we first use the multi-scale network in~\cite{li2018self} to extract multi-scale features and a two-layer MLP whose nodes are 4096 and $c$ to transform them into hash codes. Except the activation function of last layers is $tanh$ to approximate the $sign\left(\cdot\right)$ function, other layers use ReLU as activation functions. To improve generalization performance, Local Response Normalization~(LRN)~\cite{krizhevsky2012imagenet} is applied between layers of all MLPs. One should note that the application of CNN-F~\cite{chatfield2014return} and multi-scale network~\cite{li2018self} is only for illustrative purposes; any other networks can be integrated into our DSIBH as backbones of imgNet and txtNet.

As described before, semantic labels are encoded as hash codes $G^y$. To preserve semantic similarity, the loss function of labNet is:
\begin{align}\label{shi_eq7}
\underset{G^y,\theta _y}{\min}L^y&=L_{1}^{y}+\eta L_{2}^{y}\\\nonumber &=-\sum_{l,j}^{N}{\left( S_{lj}\varDelta _{lj}-\log \left( 1+e^{\varDelta _{lj}} \right) \right)}\\\nonumber&+\eta \sum_{l=1}^{N}{\left( \left\| g_{l}^{y} -f^y\left( \theta _y;y_l \right) \right\| ^2 \right)},\\\nonumber &s.t. G^y=\left\{ g_{l}^{y} \right\} _{l=1}^{N}\in \left\{ -1,1 \right\}^c
\end{align}
where $\varDelta _{lj}=f^y\left( \theta _y;\boldsymbol{y}_l \right)^Tf^y\left( \theta _y;\boldsymbol{y}_j \right)$, $f^y\left( \theta _y;\boldsymbol{y}_l \right)$ is the output of labNet for $\boldsymbol{y}_l$, $g_{l}^{y}$ is the hash codes of $f^y\left( \theta _y;\boldsymbol{y}_l \right)$ handled by $sign\left(\cdot\right)$, and $\eta$ aims to adjust the weight of loss items.

The first term of Eq.~(\ref{shi_eq7}) intends to minimize the negative log likelihood of semantic similarity with the likelihood function, which is defined as follows:
\begin{align}
p\left( \boldsymbol{S}_{lj}|f^y\left( \theta _y;\boldsymbol{y}_l \right) ,f^y\left( \theta _y;\boldsymbol{y}_j \right) \right) =\begin{cases}	\sigma \left( \varDelta _{lj} \right)&		\boldsymbol{S}_{lj}=1\\	1-\sigma \left( \varDelta _{lj} \right)&		\boldsymbol{S}_{lj}=0\\\end{cases},
\end{align}
where $\sigma \left( \varDelta _{lj} \right) =\frac{1}{1+e^{-\varDelta _{lj}}}$ is the sigmoid function. Meanwhile, the second term restricts the outputs of labNet to approximate binary as the request of hash codes.

After the optimization of labNet, the modality-invariant hash code $G^y$ is obtained. The next step is to optimize the imgNet and txtNet to generate $G^1$ and $G^2$ respectively following Eq.~(\ref{shi_eq6}). For the first item in Eq.~(\ref{shi_eq6}), DDIB interprets it as a cross-entropy loss~(i.e., Eq.~(\ref{yu_cross})). In our implement, $G^y$ is also used as class-level weight in the cross-entropy loss, which intends to make $G^1$ and $G^2$ inherent the semantic similarity of the modality-invariant hash code. Specifically, the non-redundant multi-label annotations are transformed into $N_y$-class annotations $\left\{ \bar{y}_l \right\} _{l=1}^{N_y}$, and their corresponding hash codes are regarded as the class-level weights. The weighted cross-entropy loss is formulated as:
\begin{align}\label{shi_eq8}
\min_{\theta _m} \mathcal{L}_{1}^{m}&=-\frac{1}{N}\sum_i^N{\sum_l^{N_y}{\bar{y}_l\log \left( a_{il} \right)}},\\\nonumber a_{il}&\triangleq \frac{\exp \left( \left( \bar{g}_{l}^{y} \right) ^Tg_{i}^{m} \right)}{\sum\nolimits_{l^{\prime}}^{N_y}{\exp \left( \left( \bar{g}_{l^{\prime}}^{y} \right) ^Tg_{i}^{m} \right)}},
\end{align}
where $m$ indicates the modality 1 or 2.

For the second item in Eq.~(\ref{shi_eq6}), we adopt the differentiable matrix-based R$\acute{e}$nyi's $\alpha$-order mutual information to estimate:
\begin{align}
\min_{\theta _m} \mathcal{L}_{2}^{m}=I\left( G^m;X^m \right).\label{shi_eq9}
\end{align}

For the third item in Eq.~(\ref{shi_eq6}), the $\ell _2$ loss is directly used:
\begin{align}
\min_{\theta _m} \mathcal{L}_{3}^{m}=\sum_{i=1}^N{\left( \left\| g_{i}^{y}-g_{i}^{m} \right\| ^2 \right)}.\label{shi_eq10}
\end{align}

By merging Eqs.~(\ref{shi_eq8}),~(\ref{shi_eq9}) and~(\ref{shi_eq10}) together, we obtain the loss function of imgNet~(or txtNet), formulated as the following minimization problem:
\begin{align}
\min_{\theta _m} \mathcal{L}_{}^{m}=\mathcal{L}_{1}^{m}+\beta \mathcal{L}_{2}^{m}+\gamma \mathcal{L}_{3}^{m},
\label{eq99999}\end{align}
where $\beta$ and $\gamma$ are hyper-parameters that are used to adjust the weights of loss items.

\subsection{Optimization}
The optimization of our DSIBH includes two parts: learning the modality-invariant hash code $\boldsymbol{G}^y$ and learning the hash codes $\boldsymbol{G}^1$ and $\boldsymbol{G}^{2}$ for X-ray images and radiology reports respectively. Learning $\boldsymbol{G}^y$ equals to optimize $\theta _{y+}$. For hash codes of modality $m$, $\theta _m$ needs to be optimized. The whole optimization procedure is summarized in Algorithm 1.

\begin{algorithm}[!ht]
\caption{The Optimization Procedure of DSIBH}
\textbf{Input}:  {X-ray images $\boldsymbol{X}^{1}$, radiology reports $\boldsymbol{X}^{2}$, semantic labels $\boldsymbol{Y}$, learning rates $\lambda_y, \lambda_1, \lambda_2$, and iteration numbers $T_y, T_1, T_2$.}\\
\textbf{Output}:  {Parameters $\theta _{1}$ and $\theta _{2}$ of imgNet and txtNet.}
\begin{algorithmic}[1] 
\STATE {Randomly initialize $\theta _y$, $\theta _{1}$, $\theta _{2}$ and $\boldsymbol{G}^y$.}
\REPEAT
{
\FOR{iter=1 to $T_y$}
\STATE{Update $\theta_y$ by BP algorithm:\\
$\quad\theta _y\gets \theta _y-\lambda_y \cdot \nabla _{\theta _y}L^y$}
\STATE{Update $\boldsymbol{G}^y$ by Eq.~(\ref{eq14})\;}
\ENDFOR
\FOR{iter=1 to $T_1$}
\STATE{
Update $\theta _{1}$ by BP algorithm:\\
$\quad\theta _{1}\gets \theta _{1}-\lambda_1 \cdot \nabla _{\theta _{1}}L^{1}$
}
\ENDFOR
\FOR{iter=1 to $T_2$}
\STATE{
Update $\theta _{2}$ by BP algorithm:\\
$\quad\theta _{2}\gets \theta _{2}-\lambda_2 \cdot \nabla _{\theta _{2}}L^{2}$
}
\ENDFOR
}
\UNTIL {Convergence}
\end{algorithmic}
\end{algorithm}

For $\theta _y$ of labNet, Eq.~(\ref{shi_eq7}) is derivable. Therefore, Back-propagation algorithm~(BP) with mini-batch stochastic gradient descent~(mini-batch SGD) method is applied to update it. As for $\boldsymbol{g}^y_l$, we use Eq.~(\ref{eq14}) to update:
\begin{align}
\boldsymbol{g}^y_l=sign\left( f^y\left( \theta_y;\boldsymbol{y}_l \right) \right).
\label{eq14}\end{align}

For imgNet and txtNet, we also use the BP with mini-batch SGD method to update $\theta _{1}$ and $\theta _{2}$.

\begin{table*}[!t]
	\footnotesize
	\caption{Comparison with baselines in terms of MAP on CMR-based CAD. The best results are marked with \textbf{bold}.}
	\centering
	\begin{tabular}{l c c c c c c c c}
		\hline
		\multirow{2}*{Method} & \multicolumn{4}{c}{$X\rightarrow R$} & \multicolumn{4}{c}{$R\rightarrow X$}\\ 
		\cline{2-9}
		&16 bits & 32 bits & 64 bits & 128 bits & 16 bits & 32 bits & 64 bits & 128 bits\\
		\hline
		CCA~\cite{hotelling1992relations} & 0.3468 & 0.3354 & 0.3273 & 0.3215 & 0.3483 & 0.3368 &	0.3288 & 0.3230\\
		CMSSH~\cite{bronstein2010data} & 0.4224 & 0.4020 & 0.3935 & 0.3896 & 0.3899 & 0.3967 & 0.3646 & 0.3643\\
		SCM~\cite{zhang2014large}	     & 0.4581 &	0.4648 & 0.4675 & 0.4684 & 0.4516 & 0.4574 & 0.4604 & 0.4611\\
		STMH~\cite{wang2015semantic}   & 0.3623 &	0.3927 & 0.4211 & 0.4387 & 0.3980 & 0.4183 & 0.4392 & 0.4453\\
		CMFH~\cite{ding2016large}	     & 0.3649 &	0.3673 & 0.3736 & 0.3760 & 0.4130 & 0.4156 & 0.4303 & 0.4309\\
		SePH~\cite{lin2016cross}	     & 0.4684 &	0.4776 & 0.4844 & 0.4903 & 0.4475 & 0.4555 & 0.4601 & 0.4658\\
		DCMH~\cite{jiang2017deep}	     & 0.4834 &	0.4878 & 0.4885 & 0.4839 & 0.4366 & 0.4513 & 0.4561 & 0.4830\\
		SSAH~\cite{li2018self}	     & 0.4894 &	0.4999 & 0.4787 & 0.4624 & 0.4688 & 0.4806 & 0.4832 & 0.4833\\
		EGDH~\cite{shi2019equally}     & 0.4821 & 0.5010 & 0.4996	& 0.5096 & 0.4821 & 0.4943 & 0.4982	& 0.5041\\
		\hline
		\textbf{DSIBH}	                 & \textbf{0.5001} & \textbf{0.5018} & \textbf{0.5116} & \textbf{0.5172} & \textbf{0.4898} & \textbf{0.4994} & \textbf{0.4997} & \textbf{0.5084}\\ 
		\hline
	\end{tabular}
	\label{tab1}
\end{table*}

\begin{figure*}[!h]
\centering
\includegraphics[scale=0.45]{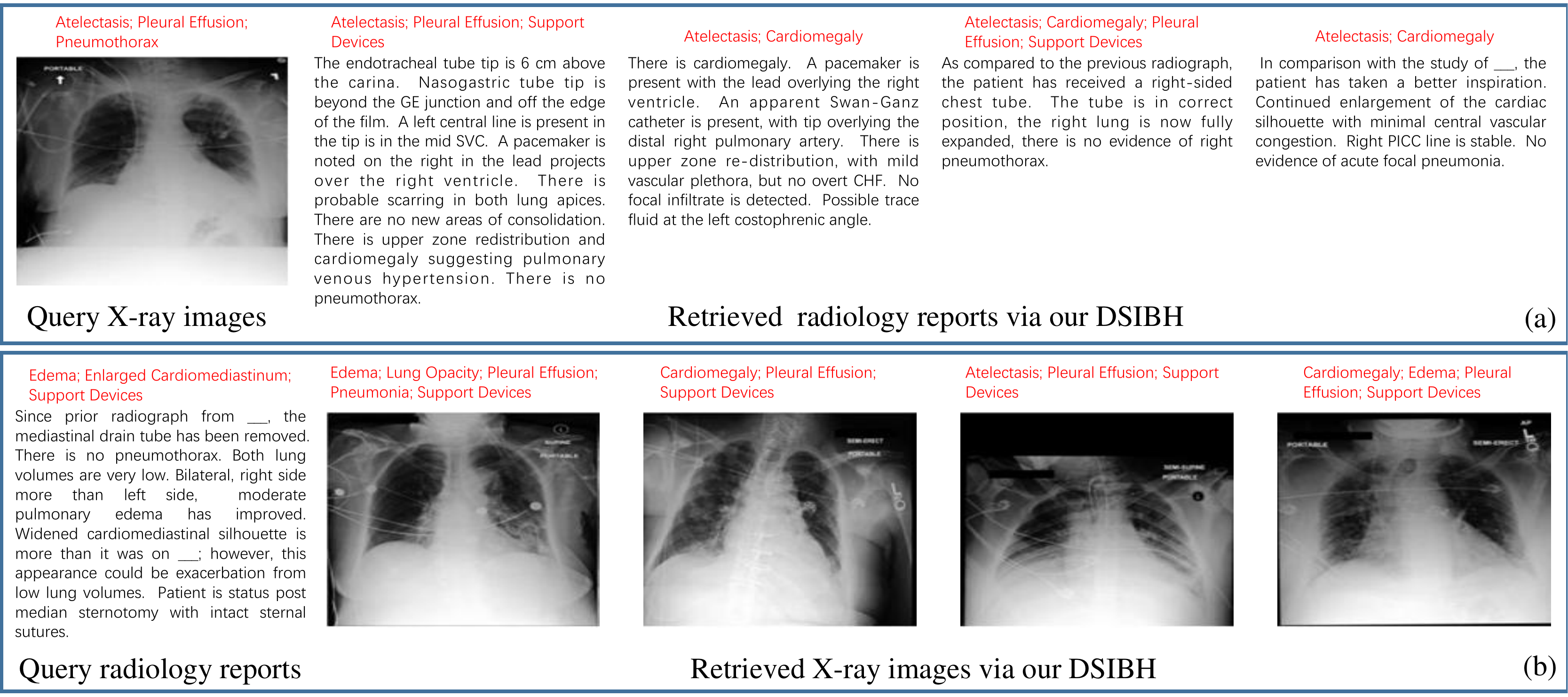}
\caption{The top 4 profiles retrieved by our DSIBH on the MIMIC-CXR dataset with 128 bits.}
\label{fig2}\end{figure*}

Once Algorithm 1 converges, the well-trained imgNet and txtNet with $sign(\cdot)$ are used to handle out-of-sample data points from modality $m$:
\begin{align}
\boldsymbol{g}_{i}^{m}=sign\left( f^{m}\left( \theta _{m};\boldsymbol{x}^{m}_i \right) \right).
\label{eq15}\end{align}

\section{Experiments}\label{sec4}

In this section, we first introduce the dataset used for assessment and specify the experimental setting. Following this, we demonstrate that the proposed DSIBH can achieve the state-of-the-art performance on CMR-based CAD. 
\subsection{Experimental setting}\label{A}
The large-scale chest X-ray and radiology report dataset MIMIC-CXR~\cite{johnson2019mimic} is used to evaluate the performance of DSIBH. Some statistics of this dataset are introduced as follows.

\textbf{MIMIC-CXR}\footnote{https://physionet.org/content/mimic-cxr/2.0.0/} consists of chest X-ray images and radiology reports sourced from the Beth Israel Deaconess Medical Center between 2011-2016. Each radiology report is associated with at least one X-ray image and annotated with a 14-dimensional label indicating the existence of pathology or lack of pathology. To evaluate the performance of CMR-based CAD, we adopt 73876 image-report pairs for assessment. During the comparison process, radiology reports are represented as bag-of-word vectors according to the top 617 most-frequent words. In the testing phase, we randomly sample 762 image-report pairs as query set and regard the rest as retrieval set. In the training phase, 14000 pairs from the retrieval set are used as training set. 

The proposed DSIBH is compared with nine state-of-the-arts in hashing-based CMR including CCA~\cite{hotelling1992relations}, CMSSH~\cite{bronstein2010data}, SCM~\cite{zhang2014large}, STMH~\cite{wang2015semantic}, CMFH~\cite{ding2016large}, SePH~\cite{lin2016cross}, DCMH~\cite{jiang2017deep}, SSAH~\cite{li2018self}, and EGDH~\cite{shi2019equally}. CCA, STMH and CMFH are unsupervised approaches that depend on data distributions, whereas the other six are supervised methods that take semantic labels into account. For fair comparison with shallow-structure-based baselines, we use the trainset of MIMIC-CXR to optimize a CNN-F network for classification and extract 4096-dimensional features to represent X-ray images.  We set $\eta =1$, $\beta =0.1$ and $\gamma=1$ for MIMIC-CXR as hyper-parameters. In the optimization phase, the batch size is set as 128 and three Adam solvers with different learning rates are applied~(i.e., $10^{-3}$ for labNet, $10^{-4.5}$ for imgNet and $10^{-3.5}$ for txtNet). 



Mean average precision~(MAP) is adopted to evaluate the performance of hashing-based CMR methods. MAP is the most widely used criteria metric to measure retrieval accuracy, which is computed as follows:
\begin{align}
MAP=\frac{1}{\left| Q \right|}\sum_{i=1}^{\left| Q \right|}{\frac{1}{r_{q_i}}}\sum_{j=1}^{R}{P_{q_i}\left( j \right) \delta _{q_i}\left( j \right)},
\label{eq16}\end{align}
where $\left| Q \right|$ indicates the number of query set, $r_{q_i}$ represents the number of correlated instances of query $q_i$ in database set, $R$ is the retrieval radius, $P_{q_i}\left( j \right)$ denotes the precision of the top $j$ retrieved sample and $\delta _{q_i}\left( j \right)$ indicates whether the $j_{th}$ returned sample is correlated with the $i_{th}$ query entity. To reflect the overall property of rankings, the size of database set is used as the retrieval radius. 

\subsection{The efficacy of DSIBH in CMR-based CAD}

CMR-based CAD stresses on two retrieval directions: using X-ray images to retrieve radiology reports~($X\rightarrow R$) and using radiology reports to retrieve X-ray images~($R\rightarrow X$). In experiments, we set bit length as 16, 32, 64 and 128 bits.

Table~\ref{tab1} reports the MAP results on the MIMIC-CXR dataset. As can be seen, unsupervised methods fail to provide reasonable retrieval results due to the neglect of semantic information. CCA performs the worst among these unsupervised methods due to the naive management of data  distribution. Compared with CCA, STMH and CMFH can achieve a better retrieval accuracy, which we argue can be attributed to the coverage of data correlation. By contrast, shallow-structure-based supervised methods including CMSSH, SCM, and SePH achieve a large performance gain over unsupervised methods by further considering semantic information to express semantic similarity with hash codes. Benefiting from the effect of nonlinear fitting ability and self-adjusting feature extraction ability, deep supervised methods including DCMH, SSAH and EGDH outperform the six shallow methods in the mass. Due to the extra consideration of superfluous information reduction, our DSIBH can achieve the best accuracy. Specifically, compared with the recently proposed deep hashing method EGDH by MAP, our DSIBH achieves average absolute increases of 0.96\%/0.47\% on the MIMIC-CXR dataset. 

Meanwhile, we also visualize the top 4 retrieved medical profiles of our DSIBH on $X\rightarrow R$ and $R\rightarrow X$ directions using the MIMIC-CXR dataset in Figure~\ref{fig2}. These results confirm our concern that DSIBH can retrieve pathology-related heterogeneous medical data again.

	
\section{Conclusion}\label{sec5}
In this paper, to preform computer-aided diagnosis~(CAD) based on the large-scale multi-modal medical data, the cross-modal retrieval~(CMR) technique based on semantic hashing is introduced. Inspired by Deep Deterministic Information Bottleneck, a novel method named Deep Supervised Information Bottleneck Hashing~(DSIBH) is designed to perform CMR-based CAD. Experiments are conducted on the large-scale medical dataset MIMIC-CXR. Compared with other state-of-the arts, our DSIBH can reduce the distraction of superfluous information, which thus strengthens the discriminability of hash codes in CMR-based CAD.
\section{Acknowledgements}
This work is partially supported by NSFC~(62101179, 61772220), Key R\&D Plan of Hubei Province~(2020BAB027) and Project of Hubei University School~(202011903000002).
\bibliography{aaai22}
\end{document}